\documentclass{article}
\usepackage{spconf,amsmath,graphicx}
\usepackage[percent]{overpic}
\usepackage{color}
\usepackage{xcolor}

\newcommand{\ie}{\textit{i.e.}} 

\title{CNN-based Preprocessing to Optimize Watershed-based Cell Segmentation in 3D Confocal Microscopy Images}

\name{{\parbox[c]{\textwidth}{\centering Dennis~Eschweiler$^{1,*}$, \qquad Thiago~V.~Spina$^{2}$, \qquad Rohan~C.~Choudhury$^{3}$, \qquad Elliot~Meyerowitz$^{4}$, \qquad Alexandre~Cunha$^{3}$, \qquad Johannes~Stegmaier$^{1,}$\sthanks{Correspondence: \texttt{dennis.eschweiler@lfb.rwth-aachen.de} or  \texttt{johannes.stegmaier@lfb.rwth-aachen.de}.}}}}
\address{$^{1}$ \small Institute of Imaging and Computer Vision, RWTH Aachen University, Aachen, Germany \\
        $^{2}$ \small Brazilian Synchrotron Light Laboratory, Brazilian Center for Research in Energy and Materials, Campinas, SP, Brazil \\
        $^{3}$ \small Center for Advanced Methods in Biological Image Analysis, Caltech, Pasadena, CA, USA \\
        $^{4}$ \small Howard Hughes Medical Institute and Division of Biology and Biological Engineering, Caltech, Pasadena, CA, USA}
\begin{document}
%
\maketitle
\begin{abstract}
The quantitative analysis of cellular membranes helps understanding developmental processes at the cellular level.
Particularly 3D microscopic image data offers valuable insights into cell dynamics, but error-free automatic segmentation remains challenging due to the huge amount of data generated and strong variations in image intensities.
In this paper, we propose a new 3D segmentation approach which combines the discriminative power of convolutional neural networks (CNNs) for preprocessing and investigates the performance of three watershed-based postprocessing strategies (WS), which are well suited to segment object shapes, even when supplied with vague seed and boundary constraints.
To leverage the full potential of the watershed algorithm, the multi-instance segmentation problem is initially interpreted as three-class semantic segmentation problem, which in turn is well-suited for the application of CNNs. Using manually annotated 3D confocal microscopy images of \emph{Arabidopsis thaliana}, we show the superior performance of the proposed method compared to the state of the art.
\end{abstract}
\section{Introduction}
\label{sec:intro}
Recent advances in fluorescent microscopy imaging technology enable observing cell dynamics at the single cell level. 
This allows analyzing a large variety of cellular developmental characteristics in detail, such as migration, differentiation, and morphogenesis.
The structures of interest are usually either cell nuclei or cell membranes, both providing complementary information of the investigated specimen.
Since cell nuclei commonly appear as separated roundish objects, they qualify for cell localization and tracking, whereas plasma membranes represent cell shape and are thus preferred for detailed morphological analysis.
Cell membranes, which are of main interest in this work, span a complex, densely connected network. 
Obtaining precise segmentations, which are essential for accurate quantitative studies, is a tedious task, especially for manual annotators. 
Varying or vanishing fluorescence intensities and increased light scattering, which are especially pronounced in structures deep inside the specimen, add an additional order of complexity to cell segmentation based on fluorescent membranes. 
Those difficulties become even more severe when advancing to a three dimensional representation, which in combination with large data sets causes manual annotations practically infeasible. \newline
\indent Existing methods to accomplish this multi-instance segmentation problem use preprocessing strategies like alternate sequential filtering \cite{Fernandez2010}, Hessian-based filtering methods \cite{Mosaliganti2012}, anisotropic diffusion filtering \cite{Pop2013} or combinations of these approaches \cite{Stegmaier2016}, to emphasize the membrane structures for further processing. The preprocessed images are then segmented using the watershed transform with seed points located in \emph{h}-minima \cite{Fernandez2010,Mosaliganti2012}, using deformable models \cite{Pop2013} or using supervoxel merging strategies 
\cite{Stegmaier2017,Funke2017}. To combine advantages of getting information from both nuclei and membranes, superposition of both experimental setups has improved generation of automated results \cite{Stegmaier2016,Zanella2010}, requiring twice the amount of data to be processed. 
Experimental constraints such as the limited amount of different fluorophores that can be expressed in an organism may prohibit the use of an additional nucleus channel.
Although deep learning models are capable of producing highly accurate segmentation results, their performance is still limited when dealing with cluttered scenes and multi-instance segmentation.
There are various approaches countering this issue by introducing customized loss functions that particularly penalize clustered segmentations \cite{Guerrero-Pena2018}.
However, seeded watershed-based algorithms proved to perform well for such multi-instance segmentation tasks.
To this end and due to the rising success of neural networks, watershed techniques have been combined with several deep learning approaches, ranging from learned watershed transforms \cite{Wolf2017} to the construction of watershed energies \cite{Bai2017}. However, these approaches are currently only suitable for 2D images and were developed for different application domains.

Although previously presented approaches perform well in outer layers of the specimen, their performance in image parts representing deep tissue regions remains limited due to trade-offs that have to be made when using global intensity-based parameters.
Moreover, membranes that are parallel to the detection plane are weakly visible.
In our previous work \cite{Stegmaier2017}, we tried to solve this problem by starting with a heavy oversegmentation followed by fusion of identified supervoxels to complete cells. This included CNN-based postprocessing steps to compensate undersegmentation errors.

While these methods improved results compared to the then state of the art, an error-free segmentation is still a challenge.
We present a new approach, formulating the problem as a 3-class segmentation, namely cell centroids, membranes and background, and combining convolutional neural networks with watershed-based segmentation techniques.
The proposed pipeline (1) improves segmentation quality in deep tissue layers, (2) does not require tedious parameter adjustments, (3) generates dense masks to separate the background and foreground regions, and (4) automatically identifies seed points for the seeded watershed method.
We validate our methods on manually annotated 3D images of \emph{Arabidopis thaliana} and show the improved performance compared to existing methods in the field.

\section{Method}
\label{sec:method}
\begin{figure}[ht] 
	\centering
	\begin{overpic}[width=0.155\textwidth]{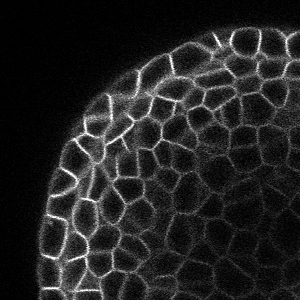}\put(1,1){\color{white} A}\end{overpic}
	\begin{overpic}[width=0.155\textwidth]{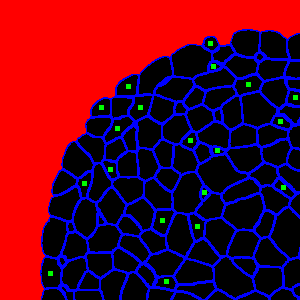}\put(1,1){\color{white} B}\end{overpic}
	\begin{overpic}[width=0.155\textwidth]{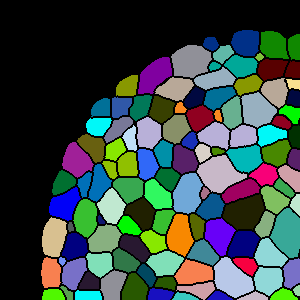}\put(1,1){\color{white} C}\end{overpic}
    \caption{(A) Cropped 2D slice from a 3D confocal image of fluorescently labeled membranes in \emph{A. thaliana}, (B) corresponding 3-class annotation with background colored in red, cell centroids in green and membranes in blue and (C) the associated multi-instance segmentation. Since this is a 2D crop from a 3D volume, centroids are only visible for segments centered at this particular slice.}
    \label{fig:sample}
\end{figure}

The proposed method combines deep learning and watershed-based segmentation into a single pipeline, exploiting the synergy of both techniques.
Therefore, the final multi-instance segmentation task (Fig.~\ref{fig:sample}C) is initially reformulated as a three-class semantic segmentation problem (Fig.~\ref{fig:sample}B).
This leverages the discriminative power of convolutional networks and in turn supplies proper seed and boundary constraints for a following seeded segmentation approach.
The three classes distinguish between cell centroids utilized for seeding, cell membranes providing detailed information about cell shapes, and background to delineate the specimen's outer surface and to reject detections in those areas.
Based on this initial setup, the proposed pipeline consists of a semantic segmentation performed by a CNN and the final multi-instance watershed segmentation.
\\\\ 
\textit{Semantic Segmentation}\\
The three-class segmentation task is performed using a modified 3D U-Net \cite{Cicek2016}, with 8 filters at the top level, 128 at the bottom level and by doubling the intermediate filter amounts at each level.
Moreover, batch-normalization is applied subsequent to each convolutional layer and input patches are normalized to zero mean and unit standard deviation.

A customized loss function is utilized for training to counteract strong imbalances between foreground and background, which is most significant for the centroid label.
The loss function is based on binary cross entropy (BCE):
\begin{equation}\label{equ_loss} 
	\begin{aligned}
    \mathcal{L}(\mathbf{y}_{t}, \mathbf{y}_{p}) = (w_{fg}\cdot \mathbf{y}_{t} + w_{bg}\cdot (1-\mathbf{y}_{t})) \cdot BCE(\mathbf{y}_{t},\mathbf{y}_{p}) \hspace*{2.3cm}
    \end{aligned}
\end{equation}
with $\mathbf{y}_{t},\mathbf{y}_{p}$ representing the ground truth and prediction masks of a particular class and $w_{fg}, w_{bg}$ constituting foreground and background weights, respectively.
Similar to the original formulation in \cite{Ronneberger2015}, foreground and background weights are added to compensate class imbalances. However, instead of a fixed weight map, the frequencies are computed on each randomly cropped training patch individually:
\begin{equation}\label{equ:weights}
	w_{fg} = \frac{1}{\sum_{i} \mathbf{y}_{t,i}}, \hspace{0.5cm} w_{bg} = \frac{1}{\sum_{i} (1-\mathbf{y}_{t,i})}
\end{equation}
with $i$ being the index of the class mask. 
The loss is calculated for each class separately and averaged outcomes are reported for further training progress. We also experimented with specimen-dependent weights introduced to the loss function, such as putting more emphasis on low-intensity boundaries or increasing the weights for membranes in deep tissue layers. However, these adaptations did not lead to higher accuracy and all reported results were performed without such adapted weights.
\\\\ 
\textit{Multi-Instance Segmentation}\\
In order to remove segmentation artifacts, the obtained probability maps are further postprocessed.
Membrane and background maps are thresholded by a fixed value of 0.5 and small objects are removed from the binary background mask. 
Centroid maps are thresholded by a value of 0.8 to obtain explicit seed locations and morphological closing with a circular structuring element of size $(5,5,5)$ further enhances explicity.
To prevent neighbouring centroids from merging, especially in deeper layers, the membrane map is subtracted from the seed map. 
To obtain the multi-instance segmentation, the binarized membrane map is converted to an Euclidean distance map \cite{Danielsson1980} with intensity minima located at cell centers.
The membrane map is added to the distance map to further enhance membrane locations and multiplied with the inverse binary background mask to exclude values exceeding the specimens boundary.
Finally, the transformed distance map and the corrected centroid map are utilized to obtain the cell instances by applying a seeded watershed algorithm~\cite{Beare2006}, again excluding segments intersecting mostly with the identified background.

In addition to the seeded watershed approach (SWS), we also adapted our previously presented supervoxel merging approach (SV) to be able to process 3D probability maps \cite{Stegmaier2017}.
We directly provide the probability maps of the membrane signal as input to the algorithm.
Subsequently, we extract supervoxels using a standard watershed algorithm by growing the watershed regions from all local minima in the image.
As the input image directly reflects the probability for each pixel of being a valid edge, we compute the average probability (\ie, the average intensity) of watershed boundaries between neighboring supervoxels and use this average probability as the merge criterion.
All supervoxel edges with an average probability below $0.5$ are considered erroneous and are thus removed.
The supervoxel merging is performed in an iterative fashion, \ie, the most unlikely edges are merged first until no valid merge candidates are left in the merge queue.
To have a baseline algorithm as well, we added a standard watershed algorithm (WS) and used all connected components below a threshold of $0.5$ as markers to grow the watershed regions.

\section{Evaluation} \label{sec:evaluation}
Experiments were evaluated using Keras\footnote{https://keras.io} with TensorFlow backend for semantic segmentation and using XPIWIT \cite{Bartschat2015} for the multi-instance segmentation.
The U-Net model was trained on 124 3D confocal image stacks of \emph{Arabidopsis thaliana} \cite{Willis2016}, which were randomly cropped to 160x160x48 patches and put into batches of size 4, restricted by memory limitations.
Validation was performed on manually annotated 3D confocal stacks of \emph{Arabidopsis thaliana}, including a total of $972$ single cells. Annotations were performed by three experts using the \mbox{SEGMENT3D} online platform \cite{Spina18}.

As comparison to the seeded watershed approach (U-Net + SWS), the native watershed (U-Net + WS) as well as the supervoxel approach (U-Net + SV) were evaluated for the multi-instance segmentation part. 
Furthermore, the results were compared to established methods in the field, namely MARS \cite{Fernandez2010} and ACME \cite{Mosaliganti2012}. The parameters for both methods were optimized using a grid search and the reported values correspond to the optimal values we obtained.
For all validation experiments aggregated Jaccard Index (JI) \cite{Kumar2017} and aggregated Dice Similarity Coefficient (DSC) were considered for assessment of the instance segmentation quality.
Boundary recall (BR) and boundary precision (BP) \cite{Neubert2012} as well as their harmonic mean (B-F1) were considered for evaluation of membrane reconstruction quality. 
Boundaries are defined to be correct, if there is a membrane within a radius of 2 voxels.

\begin{figure}[tb]
    \centering
    \begin{overpic}[width=0.139\textwidth]{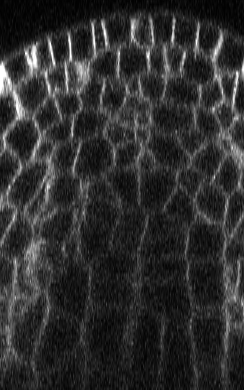}\put(1,1){\color{white}XZ}\end{overpic}
    \begin{overpic}[width=0.139\textwidth]{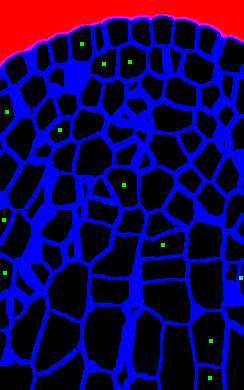}\put(-4,1){}\end{overpic}
    \begin{overpic}[width=0.139\textwidth]{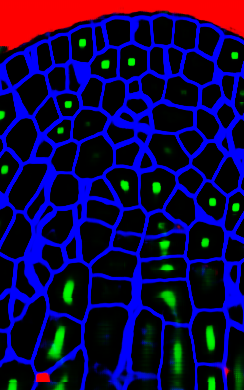}\put(-4,1){}\end{overpic}\vspace{1mm}
    \begin{overpic}[width=0.138\textwidth]{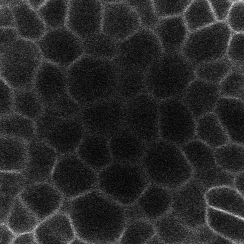}\put(1,1){\color{white}XY}\end{overpic}
    \begin{overpic}[width=0.138\textwidth]{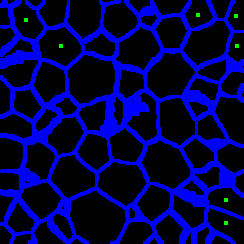}\put(-4,1){ \colorbox{white}{\color{black} Ground Truth}}\end{overpic}
    \begin{overpic}[width=0.138\textwidth]{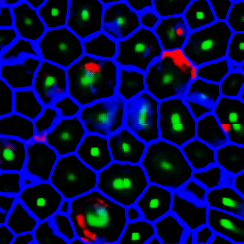}\put(-4,1){ \colorbox{white}{\color{black} Prediction}}\end{overpic}
    \caption{2D crops of a 3D confocal image of fluorescently  labeled  membranes in \textit{A. thaliana} and corresponding ground truth 3-class segmentations as well as unfiltered predictions obtained by the U-Net. The first row shows results for the xz-plane and the second row shows results for the xy-plane. Background is colored in red, cell centroids in green and membranes in blue.}
    \label{fig:3class_segm}
\end{figure}

Results for the 3-class predictions obtained by the U-Net are depicted in Fig.~\ref{fig:3class_segm}.
Note that since we present 2D crops from 3D volumes, over-segmentation of seeds in these particular slides do not necessarily indicate false detections, but rather result from less explicit centroid predictions.
Evaluation was performed for each class separately, utilizing DSC and JI for assessment of background and centroid prediction accuracy and boundary scores for assessment of membrane reconstruction quality.
Centroids are determined to be correct, if the predicted centroid is located within a cell instance and this instance has not yet been marked as detected by another centroid.
In total, background predictions achieve a JI of 0.955 and a DSC of 0.977, while centroids reach a JI of 0.755 and a DSC of 0.861.
Membrane predictions achieve a boundary recall of 0.999 and a boundary precision of 0.937, resulting in a harmonic mean of 0.967.

\begin{figure*}[htb]
    \centering
    \begin{overpic}[width=0.139\textwidth]{RAW_xz}\put(1,1){\color{white}XZ}\end{overpic}
    \begin{overpic}[width=0.139\textwidth]{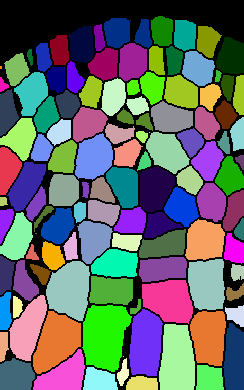}\put(-4,1){}\end{overpic}
    \begin{overpic}[width=0.139\textwidth]{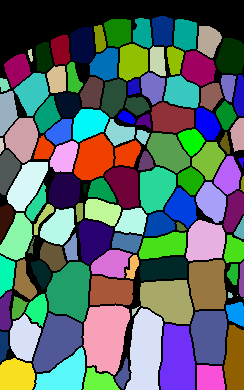}\put(-4,1){}\end{overpic}
    \begin{overpic}[width=0.139\textwidth]{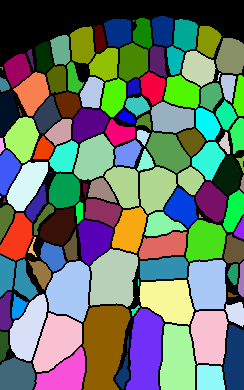}\put(-4,1){}\end{overpic}
    \begin{overpic}[width=0.139\textwidth]{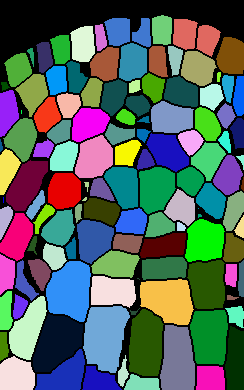}\put(-4,1){}\end{overpic}
    \begin{overpic}[width=0.139\textwidth]{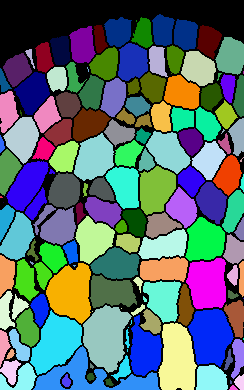}\put(-4,1){}\end{overpic}
    \begin{overpic}[width=0.139\textwidth]{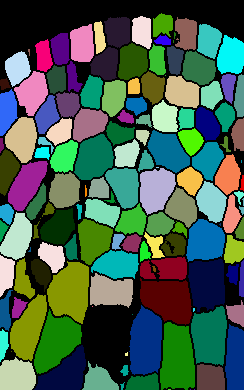}\put(-4,1){}\end{overpic}\vspace{1mm}
    \begin{overpic}[width=0.138\textwidth]{RAW_xy}\put(1,1){\color{white}XY}\end{overpic}
    \begin{overpic}[width=0.138\textwidth]{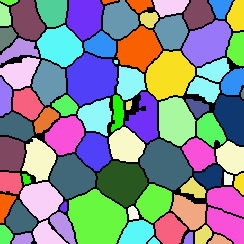}\put(-4,1){ \colorbox{white}{\color{black} Ground Truth}}\end{overpic}
    \begin{overpic}[width=0.138\textwidth]{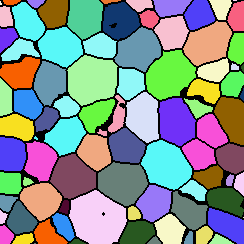}\put(-4,1){ \colorbox{white}{\color{black} U-Net+SWS}}\end{overpic}
    \begin{overpic}[width=0.138\textwidth]{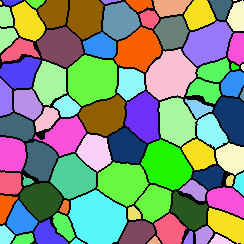}\put(-4,1){ \colorbox{white}{\color{black} U-Net+WS}}\end{overpic}
    \begin{overpic}[width=0.138\textwidth]{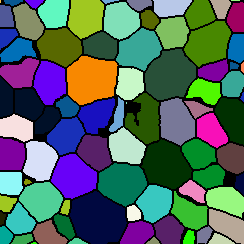}\put(-4,1){ \colorbox{white}{\color{black} U-Net+SV}}\end{overpic}
    \begin{overpic}[width=0.138\textwidth]{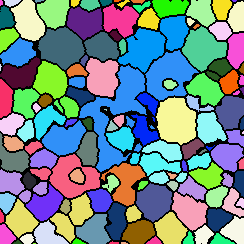}\put(-4,1){ \colorbox{white}{\color{black} MARS}}\end{overpic}
    \begin{overpic}[width=0.138\textwidth]{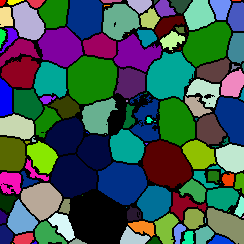}\put(-4,1){ \colorbox{white}{\color{black} ACME}}\end{overpic}
    \caption{2D crops of a 3D confocal image of fluorescently  labeled  membranes in \textit{A. thaliana} and corresponding ground truth segmentations as well as segmentations obtained from each experiment. The first row shows results for the xz-plane and the second row shows results for the xy-plane.}
    \label{fig:imgs_depth}
\end{figure*}

\begin{figure}[htb]
    \centering
    \includegraphics[width=0.47\textwidth]{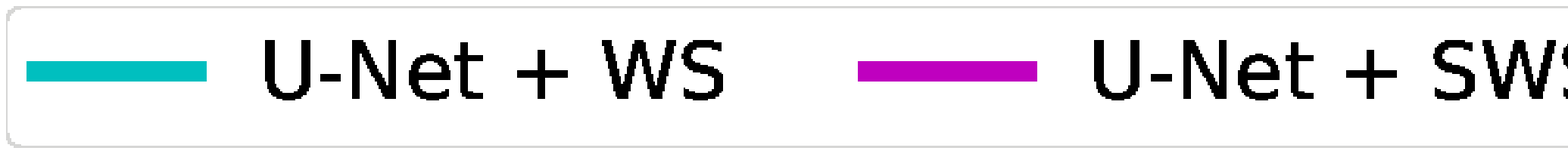}
    \includegraphics[width=0.235\textwidth]{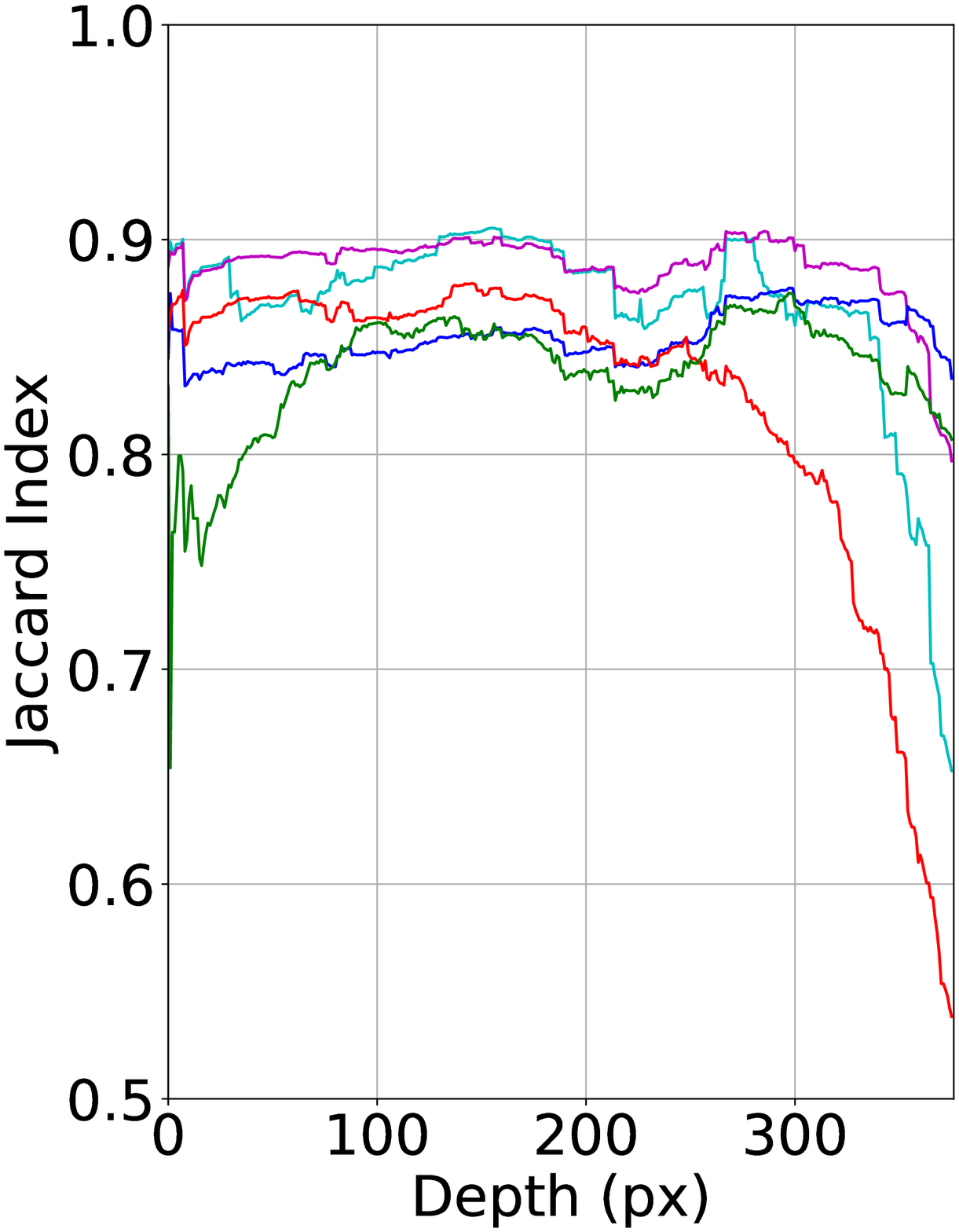}
    \includegraphics[width=0.235\textwidth]{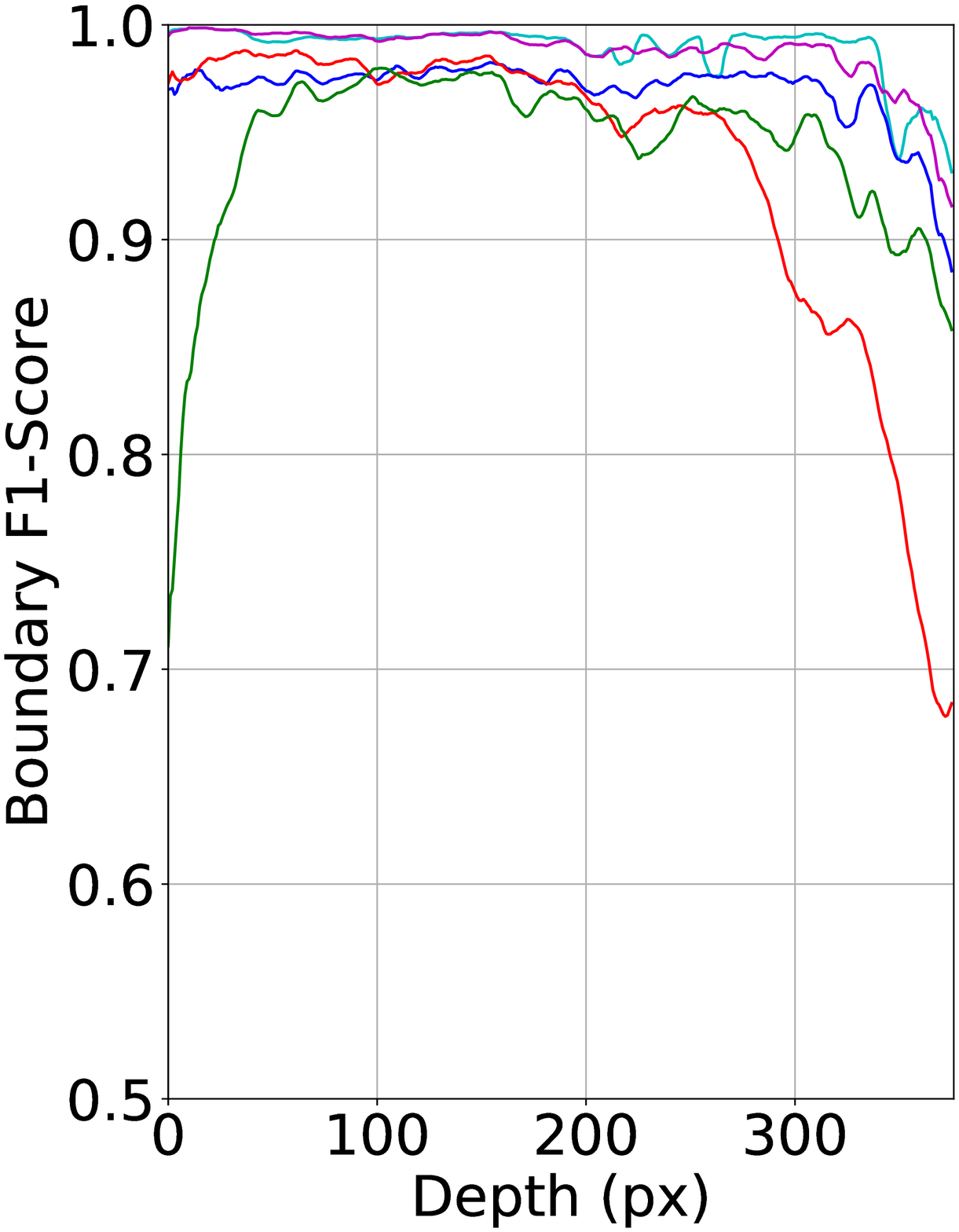}
    \caption{Aggregated Jaccard Index and boundary F1-score dependent on z-locations, plotted for each evaluated algorithm.}
    \label{fig:eval_depth}
\end{figure}

For the instance segmentation, the proposed pipeline on average achieves a JI of 0.870, a DSC of 0.931 and a boundary F1-score of 0.990 (Tab.~\ref{tab:eval_scores}).
\begin{table}[ht]
    \centering
    \begin{tabular}{c|c c c c c}
         Method & JI & DSC & BP & BR & B-F1\\
         \hline
         U-Net + SWS & \textbf{0.870} & \textbf{0.931} & 0.981 & \textbf{0.996} & \textbf{0.990}\\
         U-Net + WS & 0.843 & 0.915 & \textbf{0.989} & 0.991 & \textbf{0.990}\\
         U-Net + SV & 0.843 & 0.915 & 0.956 & 0.986 & 0.971\\
         MARS       & 0.776 & 0.874 & 0.903 & 0.977 & 0.938\\
         ACME       & 0.802 & 0.890 & 0.915 & 0.990 & 0.951\\
    \end{tabular}
    \caption{Reported scores of each algorithm, showing aggregated Jaccard Index (JI) \cite{Kumar2017} and aggregated Dice Similarity Coefficient (DSC) to assess instance segmentation quality. Additionally, boundary recall (BR) as well as boundary precision (BP) and their harmonic mean (B-F1) are presented to assess membrane reconstruction performance. Boundaries are defined to be correct, if there are membranes within a radius of 2 voxels. Bold entries indicate the best results.}
    \label{tab:eval_scores}
\end{table}
To assess segmentation quality in deeper layers, where segmentation is most challenging due to intensity variations, and to investigate dependencies between segmentation quality and layer depth, the aggregated JI and boundary F1-scores were calculated for individual layers (Fig.~\ref{fig:eval_depth}).
For computation of JI, unique labels of each ground truth z-layer were determined and 3D representation of those cells were considered for score calculation.
Depth dependent boundary scores were based on boundaries within a window size of $\pm5$ of each z-layer.
Considering global boundary scores, all U-Net-based instance segmentation approaches outperform the state of the art approaches.
While the native watershed approach (U-Net + WS) reaches global scores similar to the seeded counterpart, the benefit of utilizing seeds to be able to differentiate between cells in uncertain regions is shown in Fig.~\ref{fig:imgs_depth} and evaluated in Fig.~\ref{fig:eval_depth}.
The supervoxel approach helps to differentiate cells even when seeds fail, but due to their sensitivity towards intensity edges and potential artifacts in the cell interior, they tend to over-segment cells in high intensity regions.
Under-segmentation of the native watershed only marginally affects boundary scores (Fig.~\ref{fig:eval_depth}~right), but does have a negative impact on scores assessing the instance segmentation quality (Fig.~\ref{fig:eval_depth}~left).

\section{Conclusion}
\label{sec:conclusion}
We presented an approach for dense 3D cell segmentation, which takes advantage of predicted seed, membrane and background maps and outperforms state of the art approaches.
While seeds proved to enhance segmentation quality in deeper layers, using a supervoxel merging strategy allows to further improve segmentation in regions exposed to strong intensity fading. A combination of both techniques would potentially enhance the overall segmentation quality across varying intensity and depth levels and at the same time maintain a parameter-free pipeline. Moreover, we plan to improve the accuracy of the probability maps with additional training data of different real and simulated specimens to get closer to the ultimate goal of error-free automatic segmentation in 3D.

\bibliographystyle{IEEEbib}
\bibliography{Mendeley}

\end{document}